\documentclass[runningheads]{llncs}
\usepackage{graphicx}

\usepackage{tikz,comment,graphics,graphicx,caption,float,subcaption,booktabs,xcolor,multirow,array,color,ifthen,tabu,colortbl,dblfloatfix,url,relsize,algorithm,algorithmic,amssymb,xspace,nicefrac,microtype,amsmath,amsfonts,bm,wrapfig,makecell,enumitem}
\usepackage[accsupp]{axessibility}  %
\usepackage[pagebackref=true,breaklinks=true,colorlinks=true,bookmarks=false,citecolor=blue]{hyperref}

\begin{document}
\pagestyle{headings}
\mainmatter
\def\ECCVSubNumber{7626}  %

\title{Understanding Collapse in Non-Contrastive Siamese Representation Learning} %
%

%
%

%
%
\titlerunning{Understanding Collapse in Non-Contrastive Learning}
\author{Alexander C. Li\inst{1} \and
Alexei A. Efros\inst{2} \and
Deepak Pathak\inst{1}}
\authorrunning{A. Li et al.}
\institute{$^1$Carnegie Mellon University \quad $^2$ University of California, Berkeley}
\maketitle

\begin{abstract}
Contrastive methods have led a recent surge in the performance of self-supervised representation learning (SSL). Recent methods like BYOL or SimSiam purportedly distill these contrastive methods down to their essence, removing bells and whistles, including the negative examples, that do not contribute to downstream performance. These ``non-contrastive'' methods work surprisingly well without using negatives even though the global minimum lies at trivial collapse. We empirically analyze these non-contrastive methods and find that SimSiam is extraordinarily sensitive to dataset and model size. In particular, SimSiam representations undergo partial dimensional collapse if the model is too small relative to the dataset size. We propose a metric to measure the degree of this collapse and show that it can be used to forecast the downstream task performance without any fine-tuning or labels. We further analyze architectural design choices and their effect on the downstream performance. Finally, we demonstrate that shifting to a continual learning setting acts as a regularizer and prevents collapse, and a hybrid between continual and multi-epoch training can improve linear probe accuracy by as many as 18 percentage points using ResNet-18 on ImageNet.
Our project page is at \url{https://alexanderli.com/noncontrastive-ssl/}.

\end{abstract}

\section{Introduction}
Self-supervised representation learning (SSL) has seen steady progress in the last several years.
Recent success has been obtained via Siamese representation learning: given an input image, the neural network encoder is trained such that the feature encoding of different augmentations, aka ``views,'' of the image are close to each other.
However, trivially training such an image encoder leads to collapse, where the encoder outputs a constant representation irrespective of the input. Contrastive methods avoid this collapse by encouraging the learned representations to be far apart for views coming from very different images. There are many ways to implement this contrastive objective, either via instance discrimination, where augmentations from the same image are treated as positives and from different images as negatives~\cite{henaff2019data,he2020momentum,misra2020self,chen2020simple,bardes2021vicreg}, or by contrasting different clusters of positives~\cite{caron2018deep,caron2020unsupervised}. These approaches work well but are bottlenecked by their reliance on negatives, as it is tricky to ensure that the negative samples are not too easy to distinguish~\cite{wu2018unsupervised}.

\begin{figure}[t]
\centering
\begin{subfigure}{0.54\linewidth}
\centering
\includegraphics[width=\linewidth]{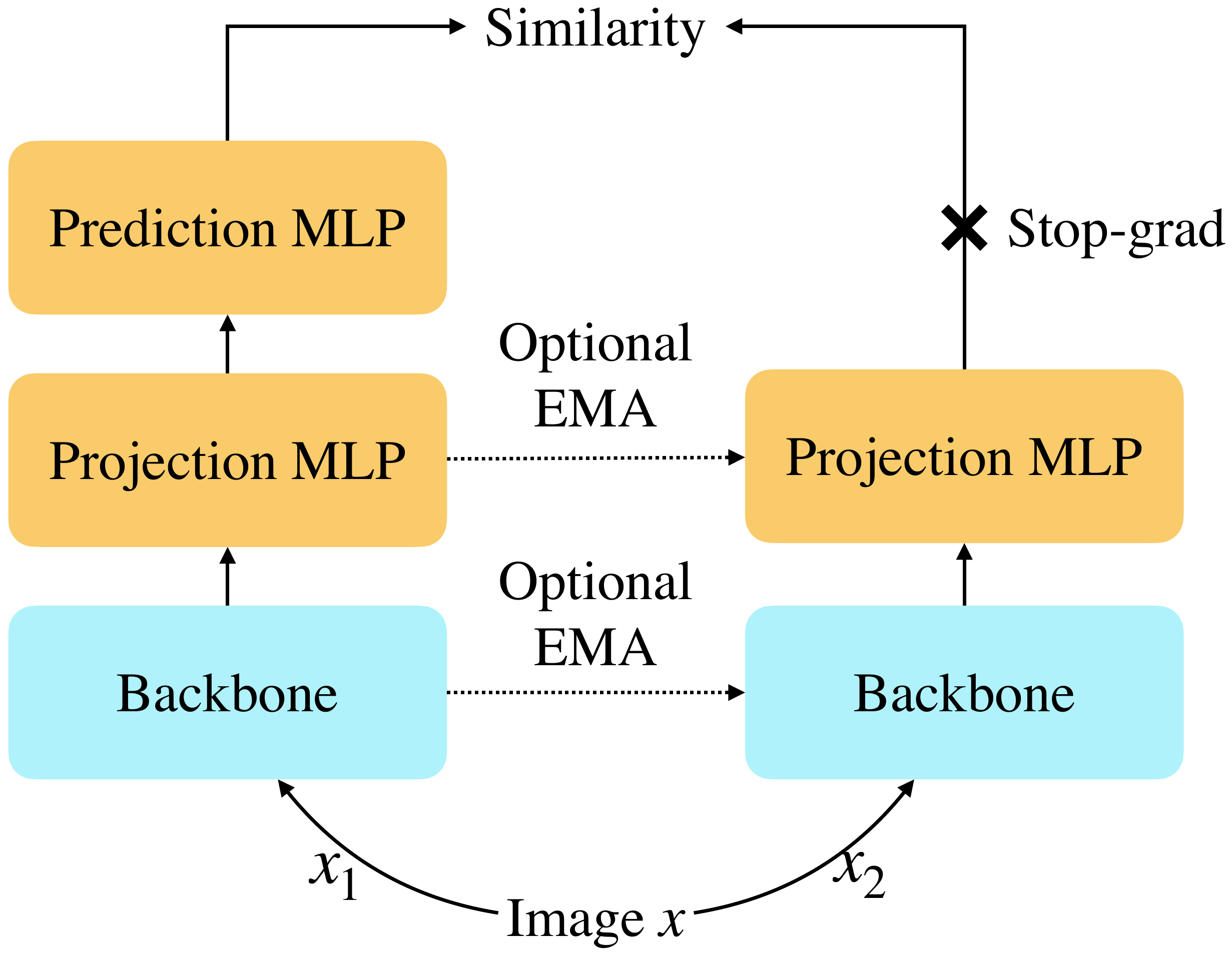}
\caption{Non-contrastive Siamese Architecture}
\end{subfigure}
\begin{subfigure}{0.44\linewidth}
\centering
\includegraphics[width=\linewidth]{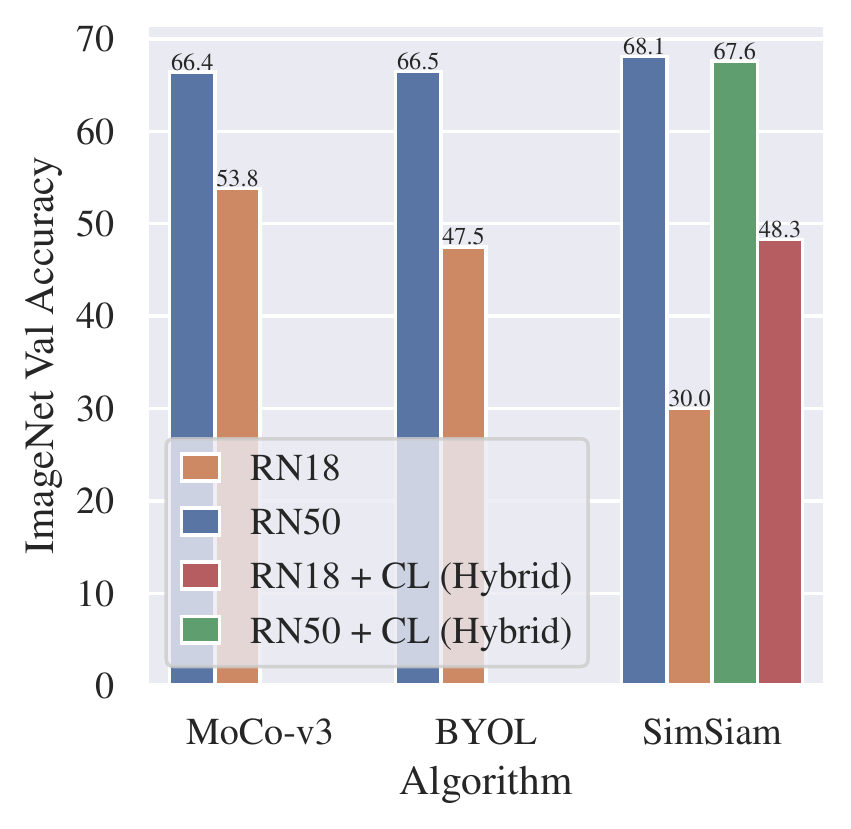}
\caption{Linear Probe Evaluation}
\end{subfigure}
\caption{\textbf{Non-contrastive methods and sensitivity to model size}. Left: given augmentations $x_1$ and $x_2$ of the same starting image, non-contrastive Siamese methods learn to use $x_1$ to predict the representation of $x_2$. SimSiam uses only the stop-grad, whereas BYOL additionally uses an exponential moving average for the second branch. Right: in contrast to methods like MoCo-v3~\cite{chen2021empirical} or BYOL~\cite{grill2020bootstrap}, SimSiam linear probe accuracy drops dramatically when the model is too small relative to the dataset complexity. We can close the performance gap and outperform BYOL on ResNet-18 by applying a hybrid of continual and multi-epoch training as discussed in Section~\ref{sec:continual}.}
\label{fig:performance_drop}
\end{figure}

Recent methods have found an alternative by removing the negatives altogether and adding architectural constraints, e.g. the momentum encoder in BYOL~\cite{richemond2020byol} or the stop gradient in SimSiam~\cite{chen2020exploring}. These ``non-contrastive'' models achieve strong results in the typical ImageNet pretraining setting, which is surprising because there is no strict constraint to prevent the aforementioned collapse. Some prior work has analyzed non-contrastive learning dynamics in a simple linear model~\cite{tian2021understanding}, but analysis of when collapse happens is still ad hoc and largely anecdotal. 

In this work, we seek to empirically understand the scenarios under which collapse occurs in these non-contrastive Siamese networks and suggest potential ways to avoid it.
Contrary to previous methods that claim that the stop-gradient, prediction head, and high predictor learning rate are enough to prevent the collapse~\cite{chen2020exploring,tian2021understanding}, we show that collapse additionally depends on the model capacity relative to the data complexity. For instance, small networks trained on large datasets are likely to collapse despite using tricks like stop-gradient or BatchNorm \cite{ioffe2015batch} on the output of the projection MLP. More importantly, the collapse need not be complete. We find that a subset of dimensions in the learned representation can collapse as well, which leads to lower than expected performance.
We define a concrete metric based on the rank of the representations to measure the degree of collapse. As expected, we show that just achieving low loss during training time does not correlate with downstream fine-tuning performance; one has to take the collapse into account as well.
 
Finally, we explore ways to prevent this collapse and find that a continual learning regime, as opposed to the dominant practice of training for multiple epochs, offers a promising alternative. An apples-to-apples comparison with the same number of total training iterations in Figure~\ref{fig:performance_drop} shows a gain of up to 18\% over vanilla SimSiam using ResNet18. We summarize our contributions below:
\begin{itemize}
    \item Contrary to previous work that claims SimSiam has no issues with collapse, we show that SimSiam performance drops significantly when the model capacity is too small relative to the data complexity. 
    \item We define a rank-based metric to measure the degree of \textit{partial dimensional collapse}, i.e., the representations contain redundant information which leads to the decline in downstream performance. 
    \item Linear regression, using our collapse metric and the SimSiam loss, can accurately predict the linear probing accuracy. This can be used to compare models without using labels or additional training time.
    \item We show that model width is more important for downstream performance than depth, even when the total number of parameters is accounted for. 
    \item We show that switching to a continual learning setting eliminates collapse and restores SimSiam accuracy by as much as 18 percentage points. 
\end{itemize}

\section{Related Work}

\paragraph{Self-Supervised Learning} Contrastive learning approaches learn a representation space where positive sample pairs are closer and negative sample pairs are driven further apart \cite{chen2020simple,he2020momentum,misra2020self,henaff2019data,oord2018representation,henaff2020data}. Often, for a given image from the dataset, a positive sample pair is constructed using an image augmentation. Negative sample pairs are generated by randomly sampling different images from the dataset. One of the drawbacks of these approaches is that training with explicit negative pairs might cause representations of very similar images to be pushed too far apart, depending on how the negatives are mined. More recent non-contrastive approaches are able to learn representations without the need for negative samples at all \cite{chen2020exploring,grill2020bootstrap}, using only image augmentations and stop gradient. 

\paragraph{Understanding Self-supervised Learning}
\cite{tian2021understanding} analyzes a surprisingly predictive linear model that represents the BYOL and SimSiam settings. However, their linear model makes no predictions about the effect of model capacity or dataset complexity. Most prior work has focused on understanding contrastive learning \cite{purushwalkam2020demystifying,jing2021understanding,tian2022deep}, but understanding of non-contrastive Siamese learning remains limited.

\paragraph{Continual Learning} Continual learning focuses on how to train models when presented with a stream of (potentially correlated) data. Typically, sticking to the multi-epoch procedure and just taking gradient steps on the data as it arrives, without any modification to the training algorithm, leads to catastrophic forgetting \cite{french1999catastrophic,kirkpatrick2017overcoming}, where model performance deteriorates on past data. A wide variety of continual learning methods have been proposed to address this problem. Regularization-based methods such as elastic weight consolidation~\cite{kirkpatrick2017overcoming} or sharpness-based regularization~\cite{deng2021flattening} seek to constrain the parameters that strongly affect the loss. Replay-based or coreset methods store summary information about previous data, such as class-based cluster centers~\cite{rebuffi2017icarl}. However, these continual learning methods have always performed worse than a model that can train on all of the data at once, i.e. multi-epoch training is preferable to continual learning. In this paper, we surprisingly draw the opposite conclusion: continual learning, without applying any fancy tricks, yields much higher downstream performance than multi-epoch training.

\section{Relative Underparameterization Causes Collapse}
\label{sec:underparameterize}

Smaller networks, such as ResNet-18 and ResNet-34 \cite{he2016deep}, are useful for a variety of reasons. 
They train faster and require less GPU memory, which is especially desirable for academics or practitioners with limited compute resources. Smaller networks also have lower latency, higher throughput, and better energy efficiency at inference time. Typically, we expect model accuracy or other metrics to fall gracefully as we reduce the size of the model. However, Figure~\ref{fig:performance_drop} shows that SimSiam performance drops significantly when using ResNet-18 instead of ResNet-50. This is unexpected, as these networks still have enough capacity to fit the SimSiam objective (BYOL, which uses the same loss function, does well). Furthermore, SimSiam with ResNet-18 has been shown~\cite{chen2021exploring} to match the performance of contrastive learning algorithms like SimCLR \cite{chen2020simple} on simple datasets like CIFAR-10 \cite{Krizhevsky09learningmultiple}. In this section, we seek to explain why smaller SimSiam models tend to have drastically lower performance on larger datasets. 

\subsection{Experimental Setup}
\label{subsec:experimental_setup}
Non-contrastive Siamese methods, e.g. BYOL and SimSiam, have the same general architecture, shown in Figure~\ref{fig:performance_drop}(a). Two views $x_1$ and $x_2$ of the same image are generated with two different augmentations, and $x_1$ is passed into the online backbone network on the left, while $x_2$ is passed into the target backbone network on the right. The backbone is typically a ResNet variant~\cite{he2016deep}. The outputs of these two backbone networks are passed into the corresponding projection MLPs, and then a prediction MLP is used to predict the projected representation of $x_2$ from the projected representation of $x_1$. SimSiam uses the same network for the online and target backbone and projection networks and uses a stop-gradient to prevent gradient signal from propagating through the second branch. BYOL also uses a stop-grad, but additionally uses an exponential moving average (EMA) to update the target backbone and projection networks. 

\paragraph{SimSiam configuration}
Unless otherwise stated, we use the same hyperparameters for each SimSiam model, from the original SimSiam ResNet-50 configuration~\cite{chen2021exploring}: we use batch size 256, which works well for ResNet-50 in the 100-epoch pretraining setting, and use SGD with learning rate 0.05, momentum 0.9, and cosine learning rate decay. Regardless of how much data we train on, or in what order the data is used, we train the model for 500 thousand gradient steps, which is equivalent to 100 epochs on 100\% of ImageNet with a batch size of 256. 
Code to reproduce our experiments is available at \href{https://github.com/alexlioralexli/noncontrastive-ssl}{{\color{blue}https://github.com/alexlioralexli/noncontrastive-ssl}}.

\paragraph{Linear probe configuration} Following the procedure in \cite{chen2021exploring}, we replace the projection and prediction heads with a fully connected layer and freeze the ResNet backbone. We perform 90 epochs of linear probing with the LARS optimizer, learning rate 0.1, and cosine learning rate decay. 
We use the standard augmentations: RandomResizedCrop, random horizontal flip, and normalization. 

\paragraph{K-nearest neighbors configuration}
We remove the projection and prediction heads, and just use the ResNet backbone to compute representations. For each image in the ImageNet-1k training and validation sets, we resize the image to 256x256, followed by a 224x224 center crop and normalization. We use cosine similarity to determine the nearest neighbors of each of the validation images, as this consistently yielded higher accuracies than the Euclidean distance.
We select the value of $k$ that maximizes the validation accuracy. The k-NN accuracy is fast to compute, requiring only about 10 minutes on an RTX 3090 GPU, compared to roughly 16 hours for linear probing. 
k-NN is also consistent and does not vary across evaluations, unlike linear probing. 

\begin{figure}[t]
    \centering
    \includegraphics[width=0.49\linewidth]{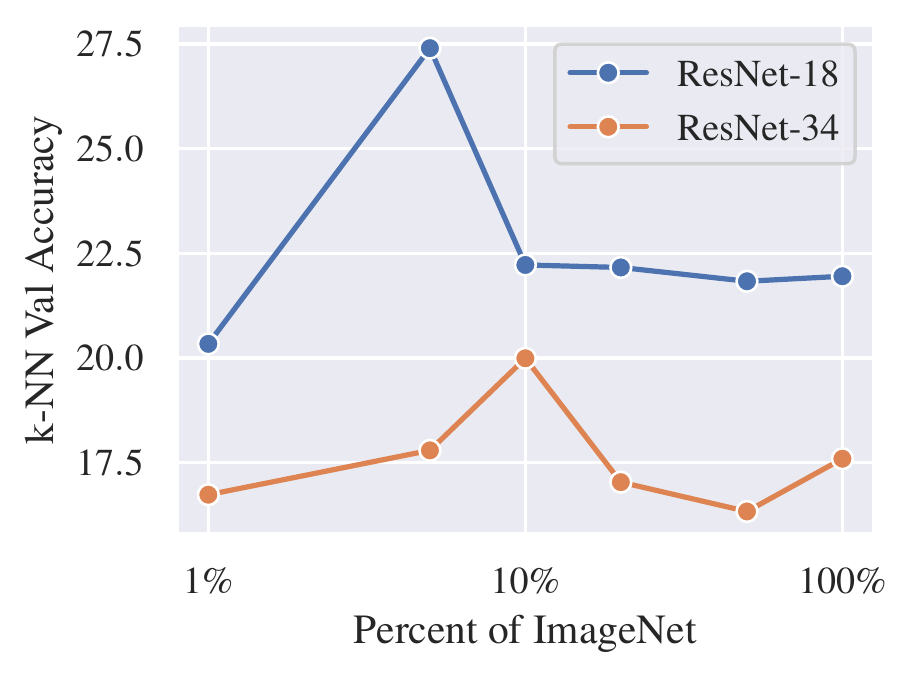}
    \includegraphics[width=0.49\linewidth]{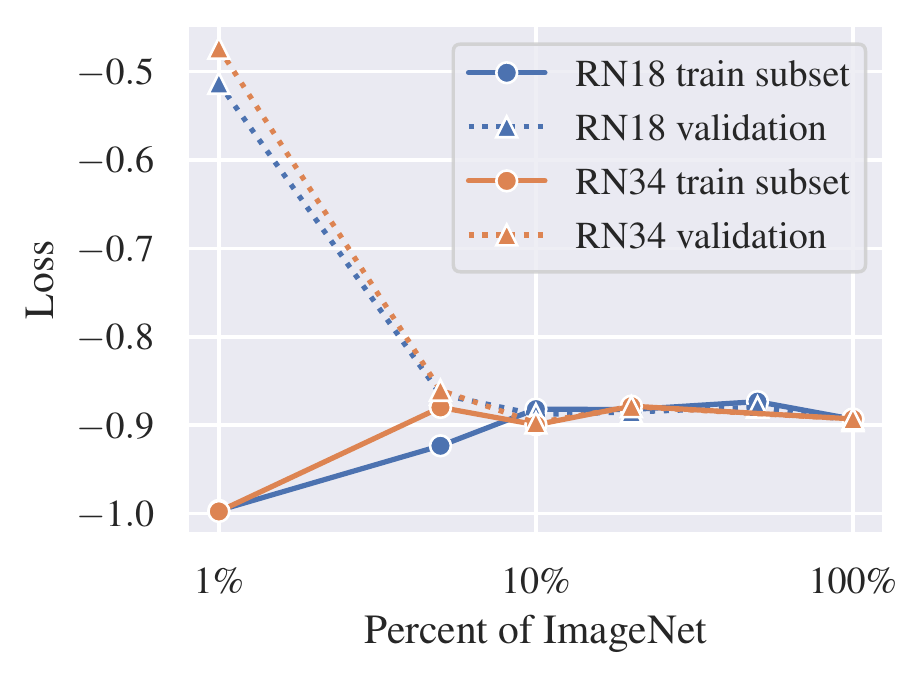}
    \caption{\textbf{SimSiam performance as a function of dataset size.} For the ResNet-18 and ResNet-34 architectures, we train 6 SimSiam models from scratch, each on a different size subset of ImageNet (1\%, 5\%, 10\%, 20\%, 50\%, and 100\%). Left: as we increase the amount of training data, the linear probing accuracy and k-NN accuracy increase until a certain model size to dataset size ratio, after which accuracy begins to fall. Right: This increase and decrease in performance is not apparent if we only look at the SimSiam loss, whether it is on the training subset or validation set.}
    \label{fig:dataset_size}
\end{figure}

\subsection{Performance Impact of Model Size Relative to Dataset Size}
\label{subsec:dataset_size}
If ResNet-18 works well with SimSiam on CIFAR-10, but does poorly on ImageNet-1k, what causes this difference? The complexity of the dataset matters as ImageNet is much more difficult to fit than CIFAR-10 (i.e. it has a larger \textit{intrinsic dimension}~\cite{li2018measuring}). Model size also matters as SimSiam + ResNet-50 is capable of SOTA performance on certain SSL benchmarks.
Hence, we hypothesize that it is actually the ratio of model capacity relative to dataset complexity that determines the SimSiam performance.
The larger and more complex the dataset, the bigger the model needs to be. 

To test our hypothesis, we perform an experiment where we train ResNet-18 and ResNet-34 SimSiam models for the same number of gradient steps but on different amounts of data from ImageNet-1k \cite{deng2009imagenet}, ranging in \{1\%, 5\%, 10\%, 20\%, 50\%, 100\%\}. By varying the size of the training set, we change how difficult it is to fit. Figure~\ref{fig:dataset_size} (left) shows the k-NN validation accuracy of these models. Both architectures have a ``sweet spot'' in the size of the training set. ResNet-18 peaks at 5\% of ImageNet-1k, whereas ResNet-34 peaks with more data at 10\%. After the peak, k-NN accuracy falls and stays relatively flat. This supports our hypothesis since more data helps up to a certain threshold. However, this increase and decrease in accuracy is not visible in any of the loss metrics shown in Figure~\ref{fig:dataset_size} (right). The ResNet-18 accuracy peaks at 5\% of ImageNet, and the ResNet-34 accuracy peaks at 10\% of ImageNet, yet the train subset's loss increases and the validation loss decreases as the model is trained on more of ImageNet.

It is worth noting that the loss on the training subset and the loss on the validation set coincide almost perfectly for every subset of at least 10\%. This indicates that the SimSiam model is not overfitting in the classical train/test sense, and theories such as ``deep double descent'' \cite{nakkiran2019deep}, which characterize the size of the effect of model size on the generalization gap between the train and test loss, do not explain the effect of varying model size or dataset size. One thing we notice is that ResNet-34 consistently performs worse than ResNet-18, which we analyze below.

\subsection{Performance Impact of Model Architecture}
\label{subsec:model_architecture}
\begin{table}[t]
    \centering
    \caption{Network width matters more than depth or number of parameters.}
    \begin{tabular}{lrrrrr}
    \toprule
        Block type & {\hspace*{4mm}}Layers & {\hspace*{4mm}}Width Multiplier &  {\hspace*{4mm}}Repr. dim. & {\hspace*{4mm}}Params & {\hspace*{4mm}}Lin. Acc. \\
    \midrule
        Basic      & 18     & 1x    & 512   & 11.7M  & 30.0\% \\
        Basic      & 34     & 1x    & 512   & 21.8M  & 16.8\% \\
        Bottleneck & 50     & 1x    & 2048  & 25.6M  & 68.1\% \\
        Bottleneck & 26     & 1x    & 2048  & 16.0M  & 61.7\% \\
        Bottleneck & 26     & 2x    & 2048  & 39.6M  & 62.6\% \\
        Basic      & 50     & 1x    & 512   & 31.9M  & 17.5\% \\
    \bottomrule
    \end{tabular}
    \label{tab:width_vs_depth}
\end{table}
In this section, we analyze what architectural components determine model capacity in non-contrastive SSL. In Table~\ref{tab:width_vs_depth}, we show the performance of various ResNet variants trained with SimSiam. The top three rows are vanilla ResNet-18, -34, and -50 models, and the last 3 correspond to new variants that mix various components. First, we find that increasing depth does not always improve performance, especially if the model is not wide enough. Increasing the depth from ResNet-18 to ResNet-34 to a depth-50 network with Basic blocks actually decreases downstream performance. We hypothesize that increased depth makes it easier for SimSiam to lose information at every layer and compute collapsed representations, since the size of the vector passed between layers is limited.

In contrast, using Bottleneck residual blocks (1x1 conv to decrease the number of channels, then 3x3 conv, then 1x1 to increase the number of channels) to increase the width of the network is much more effective. ResNet-Bottleneck-26 achieves 61.7\% linear probing accuracy with fewer parameters than even a ResNet-34. Doubling the width of that network further increases accuracy by another 0.9\%. Overall, model capacity corresponds more to width than depth. We also tried training Vision Transformers~\cite{dosovitskiy2020image} with SimSiam, but found uniformly negative results. We discuss this in Appendix \ref{subsec:vit}. 

\subsection{Performance Drops due to Partial Dimensional Collapse}
\label{subsec:dim_collapse}

\begin{figure}[t]
    \centering
    \includegraphics[width=0.325\linewidth]{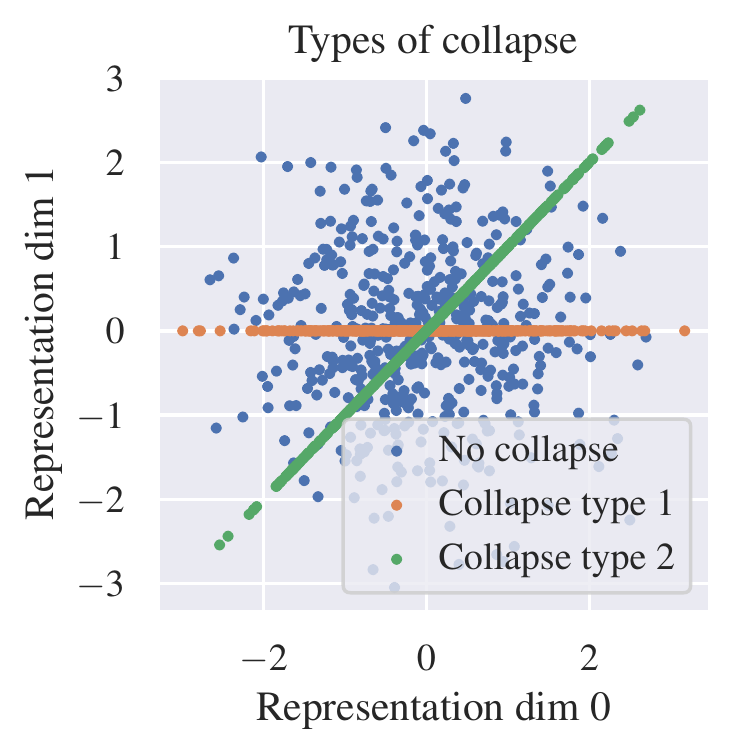}
    \includegraphics[width=0.325\linewidth]{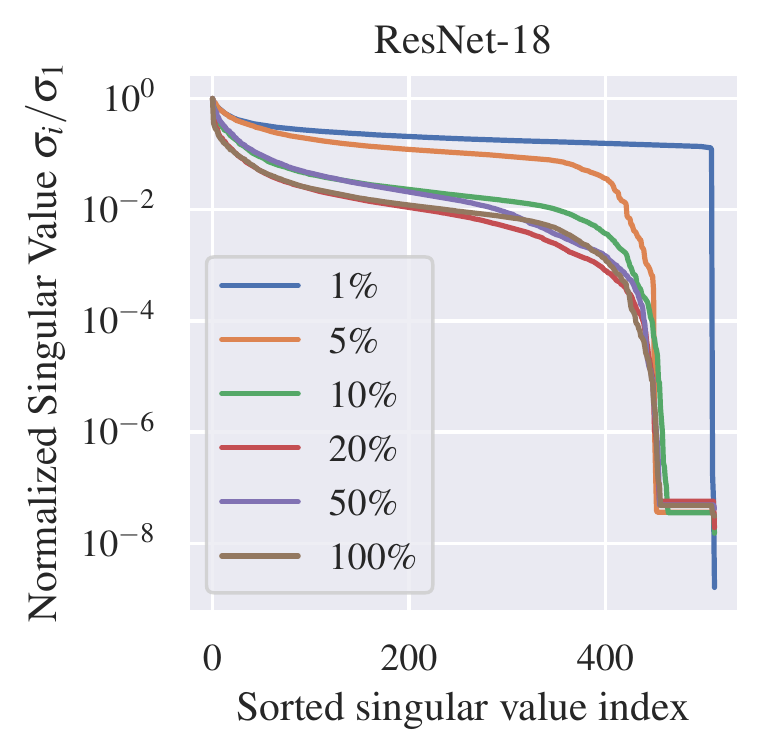}
    \includegraphics[width=0.325\linewidth]{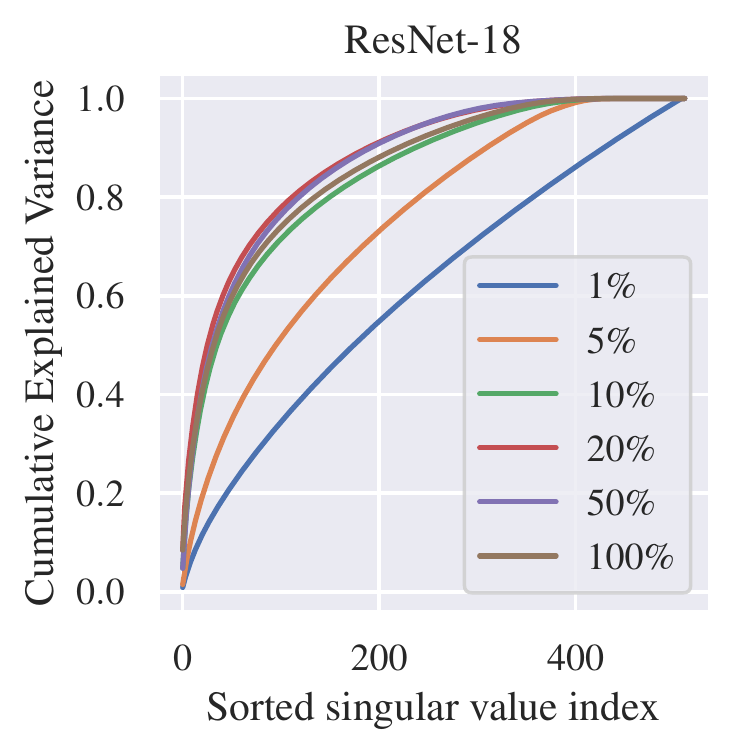}
    \caption{\textbf{Partial dimensional collapse for large subsets.} Left: the obvious form of dimensional collapse (type 1) is when a dimension collapses to a constant value. Less obvious (type 2) is when two representation dimensions covary together, i.e. one can be predicted from the other. There appears to be variation in each dimension, but the second dimension conveys no additional information. The singular values from PCA capture both kinds of collapse. Middle: we plot the singular values of representations computed by SimSiam ResNet-18 models trained on different size subsets of ImageNet. Right: the cumulative explained variance corresponding to the cumulative sum of the singular values, divided by the total. The faster this rises, the more collapse has occurred. }
    \label{fig:dimensional_collapse}
\end{figure}
We now discuss the drop in performance of ResNet18 and ResNet34 described in Section~\ref{subsec:dataset_size} and argue that it is caused by \textit{partial dimensional collapse}, where some parts of the representations either are constant across the dataset or covary with other parts of the representation. Dimensional collapse reduces the amount of information contained in the learned representations and is possible because SimSiam lacks any repulsive term to push representations apart. 

Figure~\ref{fig:dimensional_collapse} (left) shows a toy visualization of each kind of collapse. A key observation is that collapse can occur even if every representation dimension individually has high variance (see collapse type 2). This form of collapse is not captured by the collapse metric in \cite{chen2021exploring}, which for every dimension measures the standard deviation of that representation dimension across examples. In contrast, after using our model to obtain a $d$-dimensional representation for each image in the training set of $N$ samples, we perform PCA on the resulting $N \times d$ representation matrix to obtain $d$ singular values $\sigma_1 \geq \sigma_2 \geq \dots \geq \sigma_d$. More collapse should show up as smaller singular values, and PCA finds orthogonal axes that maximize variance, so it is capable of capturing both kinds of collapse.

In Figure~\ref{fig:dimensional_collapse} (middle), we examine the singular values of ResNet-18 models trained with varying amounts of data. At 5\% of ImageNet and beyond, roughly the last 80 singular values collapse to 0, and the last 300 singular values noticeably decay more when the model is trained on more data. To visualize the \textit{degree} of collapse, we look at the cumulative explained variance of the singular values: 
\begin{align}
    (\text{Cumulative explained variance})_j = \frac{\sum_{i=1}^j{\sigma_i}}{\sum_{k=1}^d \sigma_k}
\end{align}
The cumulative explained variance measures the rank of the representations and rises monotonically from 0 to 1; the more quickly it does so, the more the model has collapsed. The \{10\%, 20\%, 50\%, and 100\%\} ResNet-18 models have roughly the same explained variance curves, indicating the same high degree of collapse, which fits the fact that the k-NN accuracy flatlines for these models in Figure~\ref{fig:dataset_size}. The 1\% model exhibits no collapse at all, and the 5\% model collapses to a small degree. Despite collapsing more than the 1\% model, the 5\% ResNet-18 model has the best k-NN accuracy. We hypothesize that this is because it has much lower SimSiam loss. We explore this tradeoff in Section~\ref{subsec:predicting_acc}. 

In contrast to \cite{chen2021exploring}, which poses collapse as an ``all-or-nothing'' phenomenon that can occur when removing the stop-gradient or the prediction head, we find that collapse exists on a spectrum. Partial dimensional collapse is not unique to non-contrastive methods like SimSiam. Prior work found partial dimensional collapse in the projected embedding space for several settings, e.g., a self-similarity setup with normalization~\cite{hua2021feature} or contrastive learning with SimCLR~\cite{jing2021understanding}.

\begin{figure}[t]
    \centering
    \begin{minipage}{0.39\linewidth}
    \includegraphics[width=\linewidth]{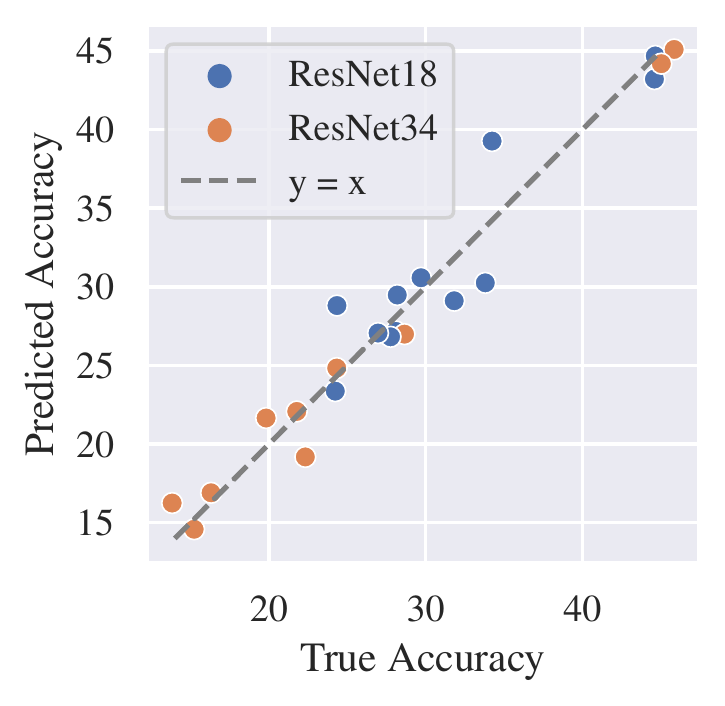}
    \end{minipage}
    \hfill
    \begin{minipage}{0.58\linewidth}
    \begin{tabular}{lrrr}
    \toprule
         & AUC Only & {\hspace*{1mm}}Loss Only & {\hspace*{1mm}}Use Both \\
    \midrule
        $R^2$              & 0.21  & 0.06  & 0.95 \\
        Pearson's $r$      & 0.46  & 0.24  & 0.98 \\
        Spearman's $\rho$  & 0.48  & 0.09  & 0.97 \\
        AUC coeff.         & -34.7 & -     & -79.5 \\
        Train loss coeff.  &    -  & -14.9 & -106.0 \\
    \bottomrule
    \end{tabular}
    \end{minipage}
    \caption{\textbf{Accuracy is predictable from loss and collapse.} We train a collection of ResNet-18 and ResNet-34 models using a variety of ImageNet subset sizes and training methods (see Section~\ref{sec:continual}). We then fit a simple linear model to predict the validation linear probing accuracy from the loss on the ImageNet validation set and the area under the explained variance curve. The simple linear model is highly predictive across both architectures, with $R^2 = 0.95$ indicating a very good linear fit. Note that either of these features alone is a poor predictor for downstream accuracy.}
    \label{fig:predicting_acc}
\end{figure}

\subsection{Predicting Performance from Collapse Metric and SimSiam Loss}
\label{subsec:predicting_acc}
There is a fundamental tradeoff between dimensional collapse and the SimSiam prediction loss. More collapse reduces the entropy in the representations and makes them more predictable, which decreases the loss, but this comes at the cost of lower representation quality. In this section, we quantify how these two properties can accurately predict model performance on a downstream task. 

We form a collection of 22 trained ResNet-18 and ResNet-34 models, consisting of the models trained on different size ImageNet subsets (from Section~\ref{subsec:dataset_size}) and models trained using different data orderings (from Section~\ref{sec:continual}). For each model, we compute the SimSiam loss on the ImageNet validation set, the validation accuracy of a linear probe trained on ImageNet, and a metric that measures the degree of dimensional collapse in the representations. Following the same PCA procedure as Section~\ref{subsec:dim_collapse}, we compute the $d$ singular values $\sigma_1, \dots, \sigma_d$ of the representations of the full training set. Our collapse metric corresponds to the area under the cumulative explained variance of the singular values: 
\begin{align}
    \text{AUC} &= \frac{\frac{1}{d}\sum_{i=1}^{d} \sum_{j=1}^i \sigma_j}{\sum_{k=1}^d \sigma_k}
\end{align}
The AUC can range from 0.5 to 1, and larger AUC values reflect more collapse. $\text{AUC}=0.5$ means all of the singular values are identical, which indicates that no collapse is occurring. $\text{AUC}=1$ means that the last $d-1$ singular values are 0, which indicates severe dimensional collapse.

Using values computed from our collection of 22 SimSiam models, we fit a linear model to predict the validation accuracy from the loss and AUC. Figure~\ref{fig:predicting_acc} shows that this linear model is highly accurate, with $R^2=0.950$. Strongly negative coefficients on the loss and AUC make sense: lower loss and less collapse result in more useful features. Figure~\ref{fig:predicting_acc} (left) shows that this linear fit works for both ResNet-18 and ResNet-34, with accurate predictions for most models. Note that the loss and AUC are only jointly predictive; using only loss or only AUC poorly predicts the downstream accuracy. 

This finding has two implications. First, it allows estimating models' relative performance without using any labels. We can simply compute the SimSiam loss on the validation set, as well as the singular values of the representations on the training set. If one model dominates the other in both metrics, i.e. it has lower loss and less collapse, then it is obviously better. Otherwise, we weight the loss and collapse metric by this linear formula to choose which model to fine-tune for a desired downstream task. In addition to not requiring labels, this procedure also eliminates the need to fine-tune multiple models to see which is best.  Second, this finding means that we can improve SimSiam performance either by decreasing the loss or by reducing collapse. Section~\ref{sec:continual} presents methods that focus on the latter.

\section{Continual Training Prevents Collapse}
\label{sec:continual}
Section~\ref{sec:underparameterize} showed that SimSiam models tend to collapse if the training set is too large relative to the size of the network. At first glance, it seems that the only solutions are to use a larger model, which requires more time and compute, or to switch to a different self-supervised learning algorithm, which may itself have its own disadvantages (e.g. SimCLR \cite{chen2020simple} requires big batch sizes, and BYOL requires twice as many forward passes per update). However, in this section, we find that SimSiam's dimensional collapse has a simple solution that requires no change to the architecture, loss function, or hyperparameters. Motivated by the observation in Figure~\ref{fig:dimensional_collapse} that shows that collapse does not occur in models trained on small subsets of the data, we simply partition the training set into small subsets and train on them in sequence. We show that changing the data order by switching to a \textit{continual learning setup} is surprisingly highly effective at preventing dimensional collapse in small models. 

\begin{figure}[t]
    \centering
    \includegraphics[width=\linewidth]{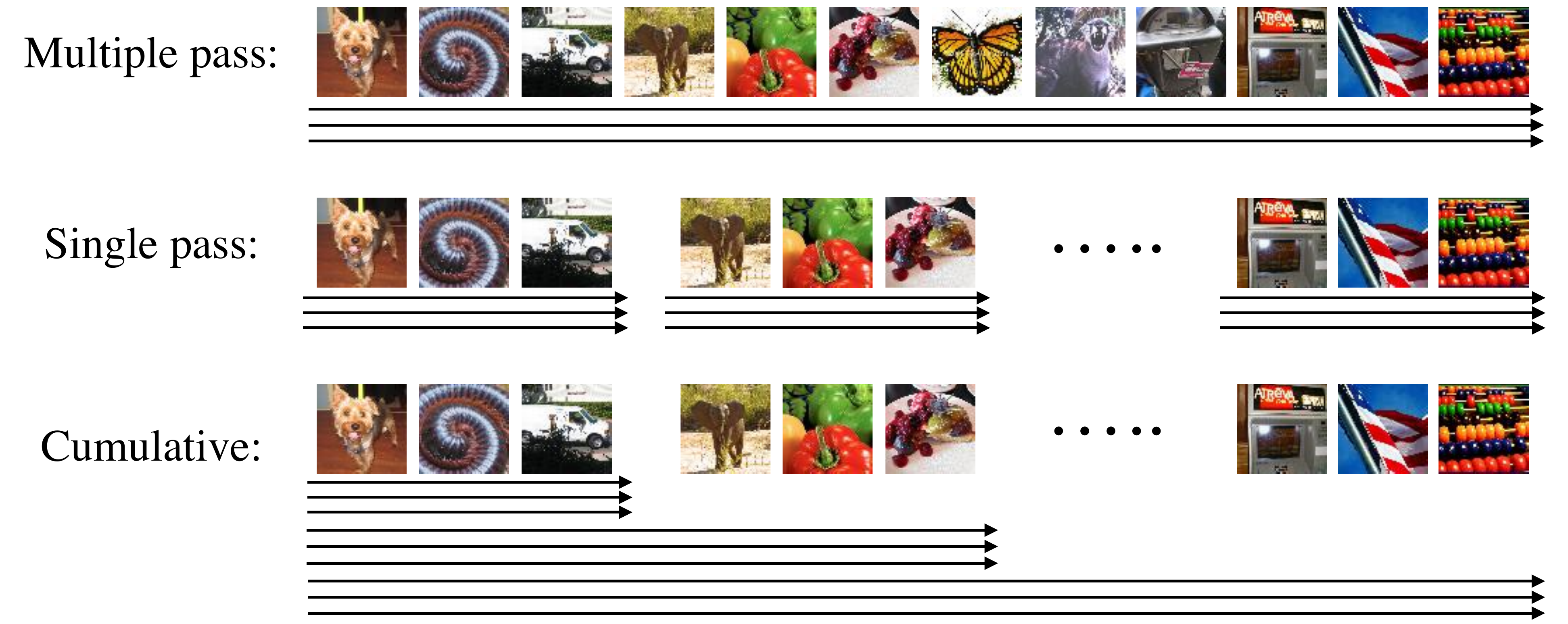}
    \caption{\textbf{Illustration of each data ordering method.} ``Multiple pass'' training consists of making $E$ passes over the training set, so each data point is seen infrequently but uniformly across training. ``Single pass'' is a form of continual learning that splits the dataset into chunks. It trains intensely on a chunk, then throws it away and moves on to the next chunk. ``Cumulative'' also splits the dataset into chunks, but never throws away data. It accumulates data as it arrives, and begins to approximate ``multiple pass'' training towards the end of training.}
    \label{fig:method_illustration}
\end{figure}

We change only the order in which data is presented to the model; we use the same architecture, loss function, number of gradient steps, and hyperparameters as before. We compare the three possible data orderings illustrated in Figure~\ref{fig:method_illustration}: 
\begin{enumerate}
    \item Multiple pass: this is the standard multi-epoch training procedure. The model is trained for $E$ epochs, and each training image is used once each epoch. 
    \item Single pass: this is the \textit{data-incremental} setting from continual learning. The data is randomly shuffled and partitioned into $C$ chunks, which ``arrive'' one after another. When a chunk arrives, we take $N$ stochastic gradient steps, after which we \textit{throw away the chunk and stop using its images.} By default, we set the number of chunks to $C=100$.
    \item Cumulative: this is akin to the incremental learning setting with a replay buffer that is large enough to hold the entire training set. When a chunk arrives, we add its images to the replay buffer and then take $N$ stochastic gradient steps. As more chunks arrive, the total size of the replay buffer increases and approaches the ``multiple pass'' setting towards the end. 
\end{enumerate}

\begin{table}[t]
    \centering
    \caption{ImageNet top-1 linear probing validation accuracy for different SimSiam training methods. We show the mean and standard deviation over 3 random seeds.}
    \begin{tabular}{lr@{\hspace*{4mm}}r@{\hspace*{4mm}}r}
    \toprule
        Training method & ResNet-18 & ResNet-34 & ResNet-50 \\
    \midrule
        Mutiple pass & $30.0 \pm 1.8$                  & $16.8 \pm 3.2$ & \textbf{68.1} \\
        Cumulative               & $33.0 \pm 1.9$ & $22.2 \pm 2.3$ & 67.7 \\
        Single pass                           & $44.5 \pm 0.8$ & $45.0  \pm 1.1$ & 55.9 \\
        Hybrid (switch at 40) & $\mathbf{48.3} \pm 0.7$ & $\mathbf{50.3} \pm 0.6$ & 67.6 \\
    \bottomrule
    \end{tabular}
    \label{tab:continual_comparison}
\end{table}

\subsection{Results}
\begin{figure}[t!]
    \centering
    \includegraphics[width=0.32\linewidth]{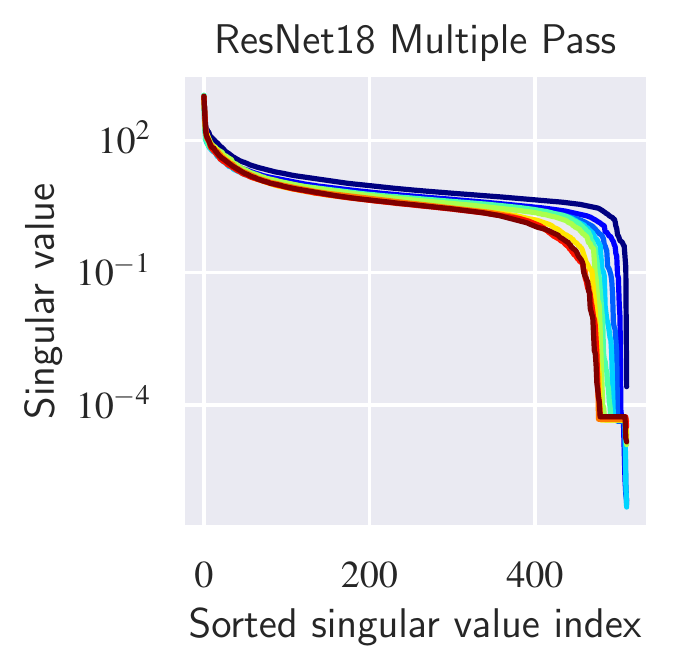}
    \includegraphics[width=0.32\linewidth]{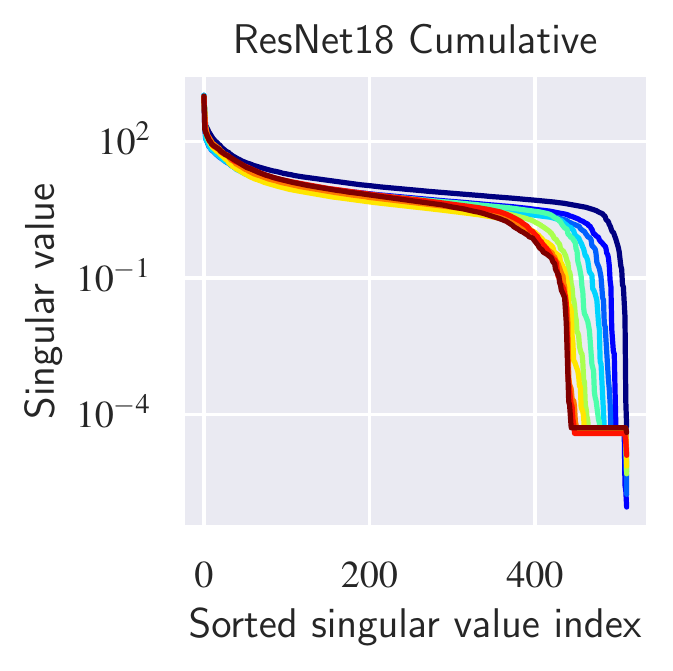}
    \includegraphics[width=0.32\linewidth]{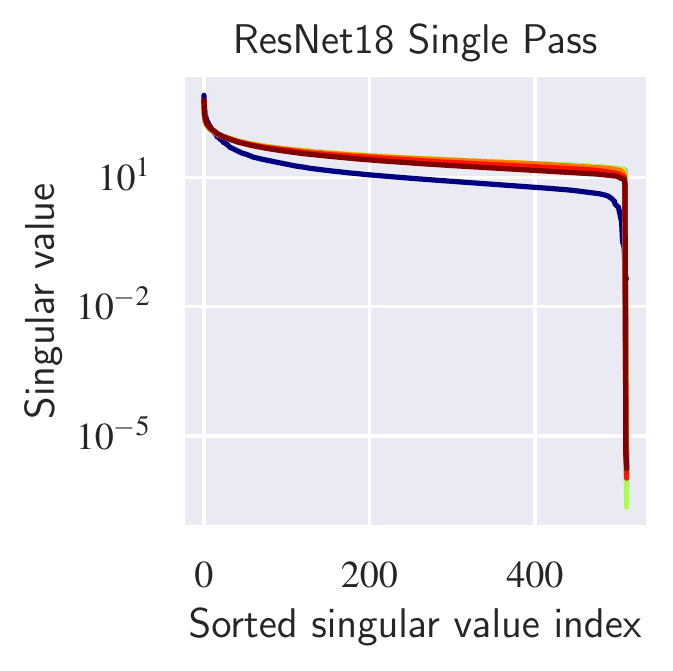}
    \includegraphics[width=0.32\linewidth]{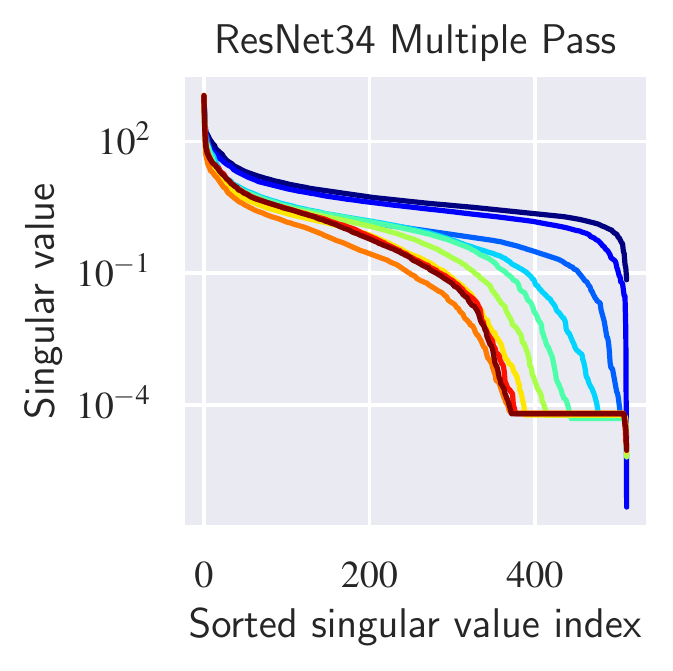}
    \includegraphics[width=0.32\linewidth]{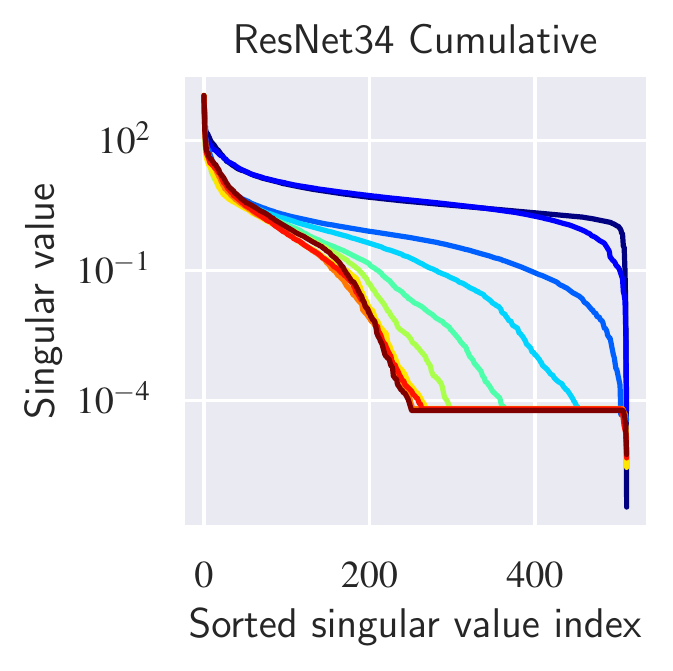}
    \includegraphics[width=0.32\linewidth]{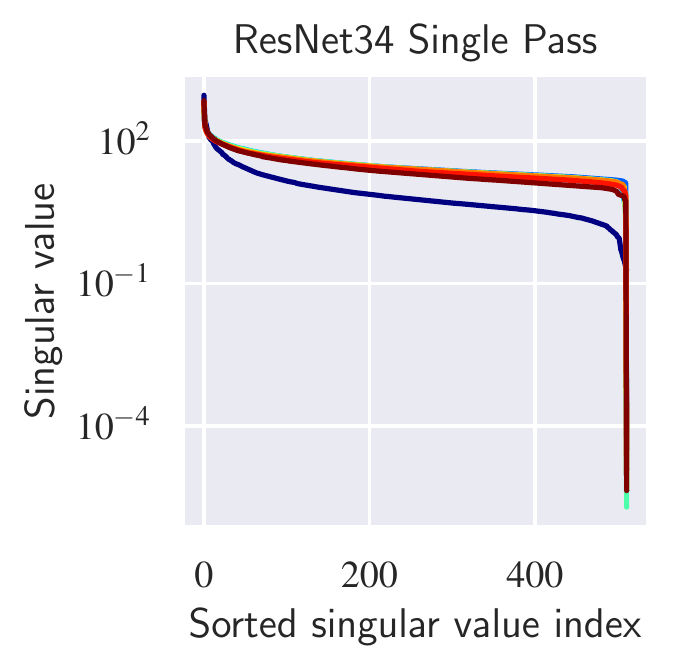}
    \includegraphics[width=0.32\linewidth]{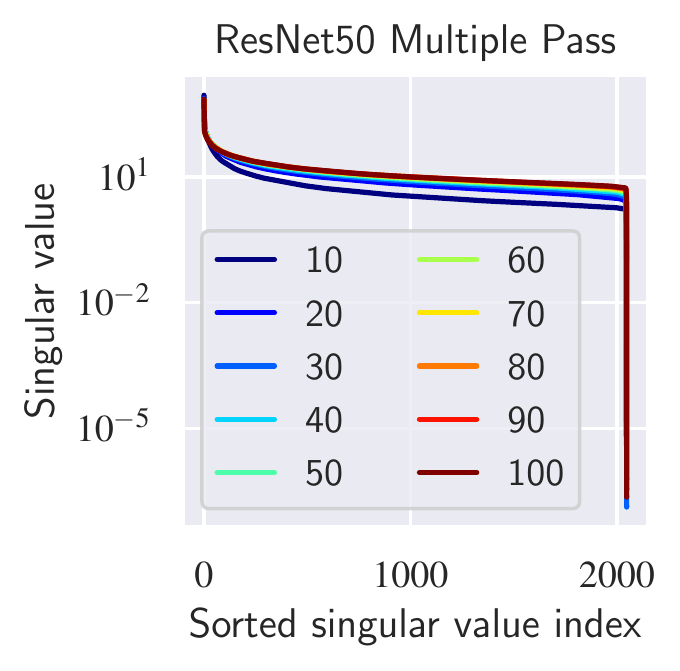}
    \includegraphics[width=0.32\linewidth]{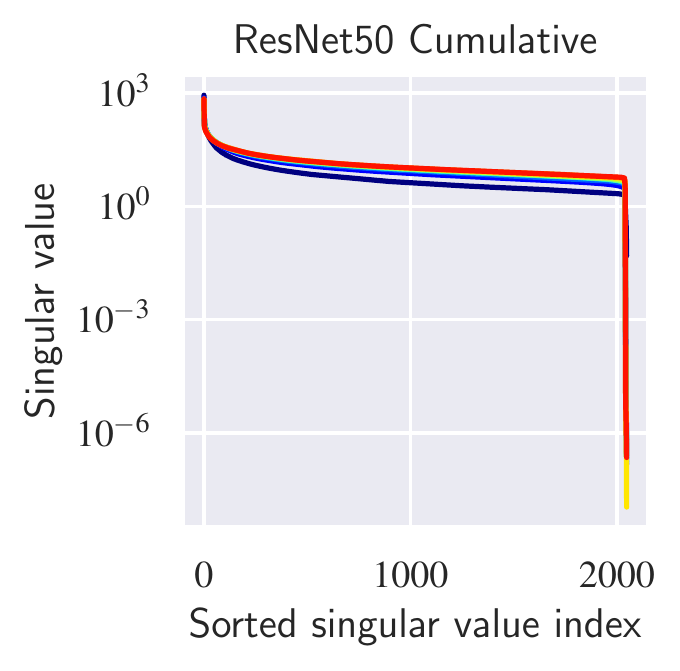}
    \includegraphics[width=0.32\linewidth]{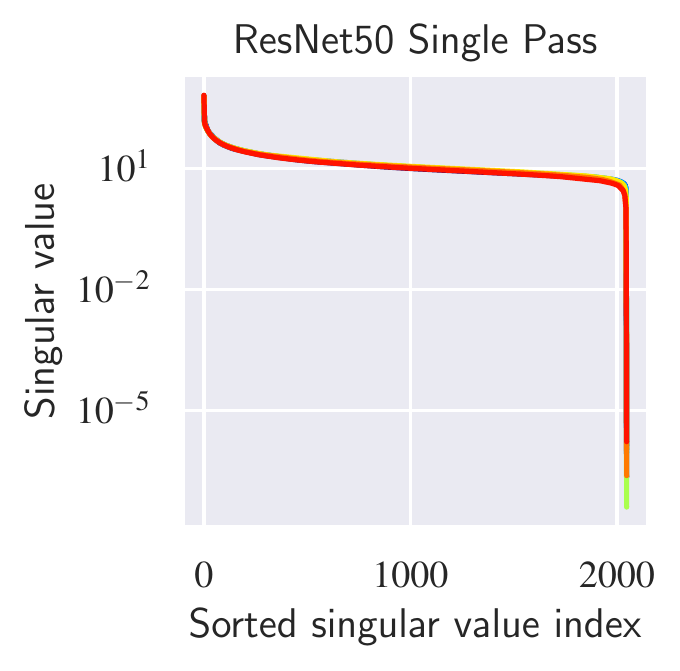}
    \caption{\textbf{Evolution of dimensional collapse across training}. Each line shows the singular values corresponding to an intermediate training checkpoint. For ``multiple passes'' with both ResNet-18 and ResNet-34, more singular values collapse towards 0, indicating that dimensional collapse is happening. In contrast, the ``single pass'' strategy avoids collapse and in fact \textit{increases} the singular values across training. ResNet-50 does not collapse even in the multi-epoch setting, so continual training does not help here.}
    \label{fig:continual_collapse}
\end{figure}

Table~\ref{tab:continual_comparison} shows the result when using these 3 data orderings to train ResNet-18, ResNet-34, and ResNet-50 architectures. Continual training (``single pass'') \textit{improves validation accuracy by 14.5 percentage points (ResNet-18) and 28.2 percentage points (ResNet-34)} over the ``multiple pass'' baseline, but leads to a 12.2\% drop for ResNet-50. Figure~\ref{fig:continual_collapse} shows that continual training helps ResNet-18 and ResNet-34 by preventing collapse, but ResNet-50 with ``multiple pass'' training does not collapse in the first place since ResNet-50 is sufficiently big. Thus, continual training for ResNet-50 introduces catastrophic forgetting \cite{french1999catastrophic,kirkpatrick2017overcoming}, making it more difficult to achieve low SimSiam loss. 

Accumulating data seems natural since it acts as a curriculum and gives the model time to fit each new image, but it only yields minor improvements. It results in a 3-6 percentage point improvement over multi-epoch training for ResNet-18 and ResNet-34 and is close to matching multi-epoch training for ResNet-50. However, this is still far behind ``current,'' and Figure~\ref{fig:continual_collapse} shows that ``cumulative'' training does not significantly help prevent collapse. 

\begin{figure}[t]
    \centering
    \begin{minipage}{0.44\linewidth}
    \includegraphics[width=\linewidth]{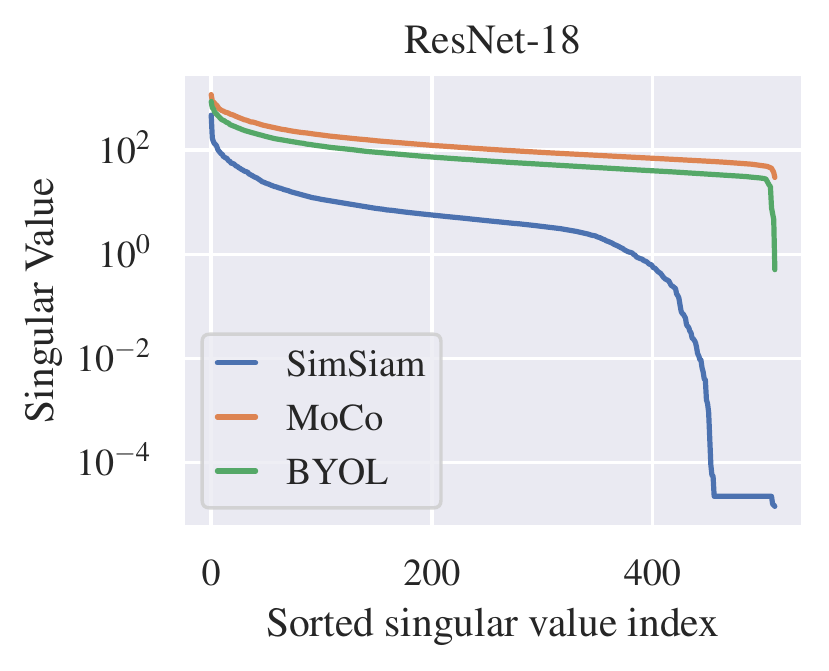}
    \end{minipage}
    \hfill
    \begin{minipage}{0.48\linewidth}
    \begin{tabular}{l@{\hspace*{6mm}}r@{\hspace*{4mm}}r}
    \toprule
        Training method & MoCo-v3 & BYOL  \\
    \midrule
        Multiple pass & \textbf{53.8} & \textbf{47.6} \\
        Single pass   & 48.9          & 44.2 \\
    \bottomrule
    \end{tabular}
    \end{minipage}
    \caption{Singular value analysis shows that MoCo and BYOL, unlike SiamSiam, do not collapse with small models. Thus, single pass training does not improve ResNet-18 ImageNet linear probe accuracy.}
    \label{fig:no_benefit}
\end{figure}

Finally, continual training does not help with other SSL algorithms, such as MoCo-v3 \cite{chen2021empirical} or BYOL \cite{grill2020bootstrap}, as shown in Figure~\ref{fig:no_benefit}. This makes sense since the plot of their singular values shows that these methods do not collapse. This indicates that the exponential moving average (EMA) in BYOL prevents collapse and is actually crucial outside of the standard ImageNet-1k pretraining benchmark with ResNet-50. Note that using EMA introduces two additional costs at training time. First, the EMA itself requires maintaining two sets of weights that are constantly updated, which uses more GPU memory. 
Second, calculating the symmetrized loss of 2 sets of views requires twice as many forward passes, since the target network has different parameters than the online network. 

In contrast, our proposed data ordering methods allow us to keep the efficiency of SimSiam with no additional overhead. Furthermore, switching to the continual setting offers its own advantages. Continual methods are faster to train, especially with limited resources, since fitting the entire chunk in memory allows for faster access than epoch-wise retrieval from disk or an NFS. Continual methods are also well suited for real-life applications, where data often arrives in a stream~\cite{purushwalkam2022challenges}. 
Algorithmic improvements in continual training for SSL will likely transfer well to these practical settings. 

\subsection{Hybrid of Continual and Multi-epoch Training}
\begin{figure}[t!]
    \centering
    \includegraphics[width=0.49\linewidth]{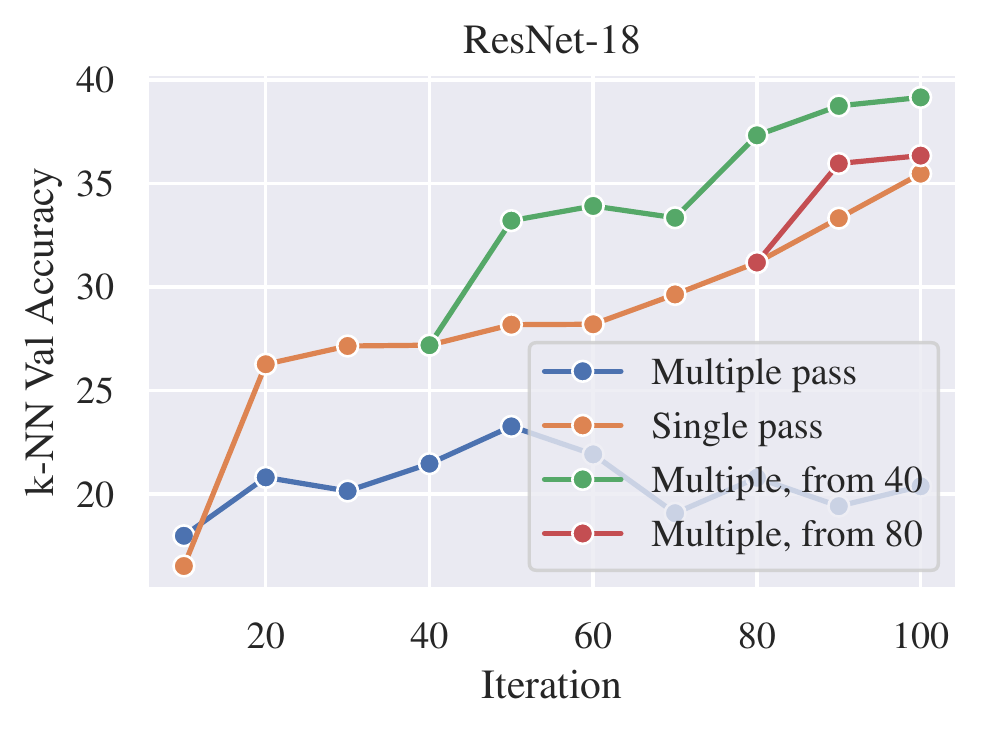}
    \includegraphics[width=0.49\linewidth]{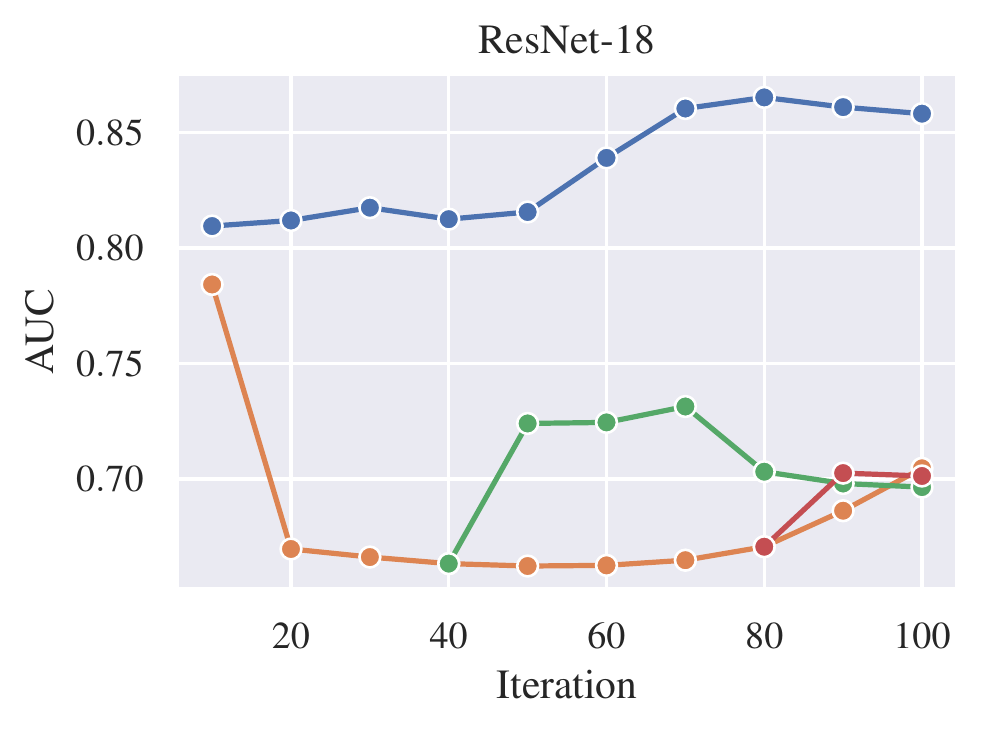}
    \includegraphics[width=0.49\linewidth]{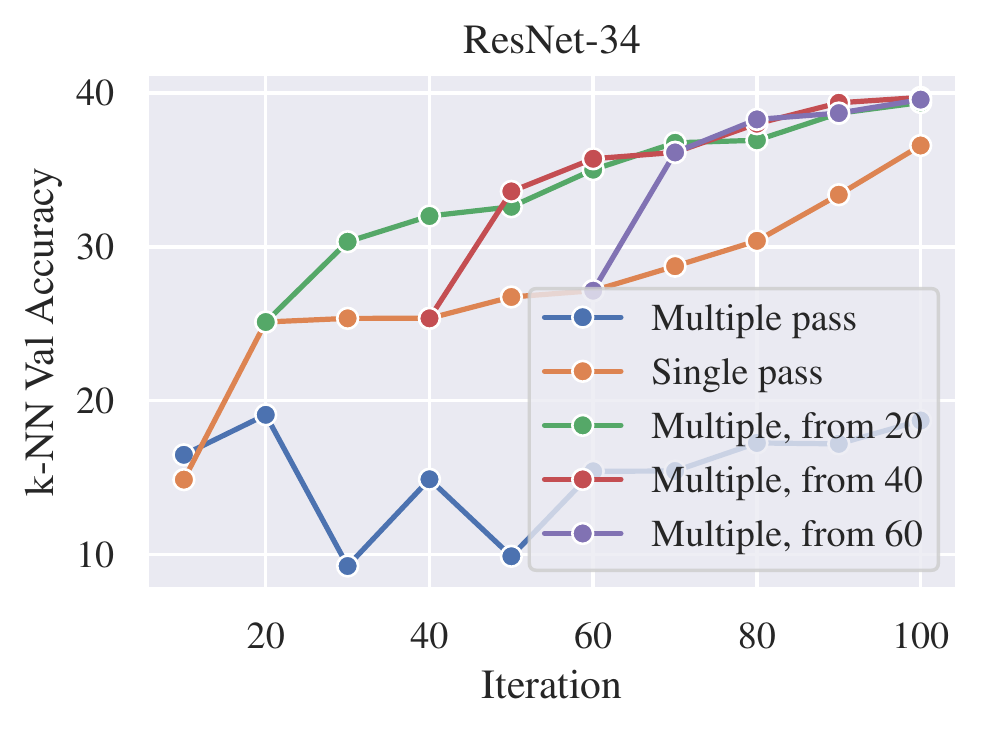}
    \includegraphics[width=0.49\linewidth]{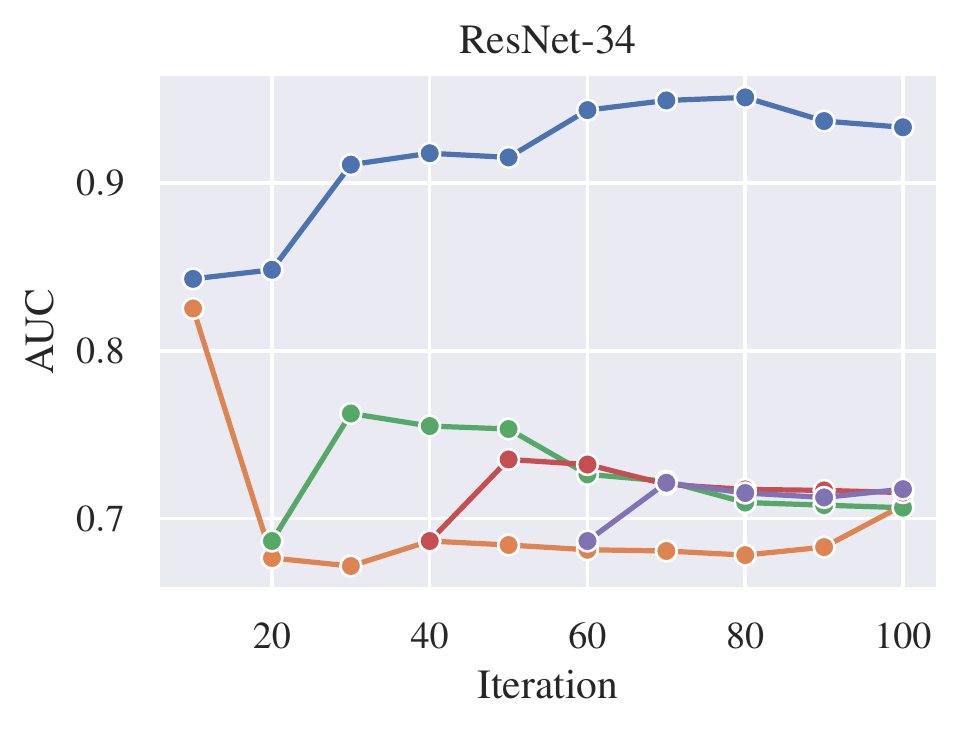}
    \caption{\textbf{Hybrid of Continual and Multi-epoch Training Improves Performance} We take an intermediate ``single pass`` checkpoint and fine-tune it with multi-epoch training. Our partial collapse metric (AUC) shows that multi-epoch fine-tuning tries to collapse immediately, but the initialization is good enough that it recovers. }
    \label{fig:across_training}
\end{figure}

Even though the pseudo-curriculum implemented in the ``cumulative'' data ordering did not improve performance, warming up the network with one method and then finishing training with another could be useful for simultaneously (a) preventing collapse while (b) driving the SimSiam loss lower. 
We experiment with doing continual training for the first part of training, then switching to multi-epoch training for the rest of training. 
Figure~\ref{fig:across_training} shows how the k-NN accuracy and AUC collapse metric change over the course of training. 
Continual training for 40 chunks followed by multi-epoch training achieves the highest k-NN validation accuracy, but only works because of the continual training in the beginning.
Right after switching training methods, the model begins to collapse, but the initialization from continual training gives the model more room to recover. 
Table~\ref{tab:continual_comparison} shows that this hybrid method achieves 48.3\% linear probing accuracy with ResNet-18, \textit{18.3 percentage points better than the multi-epoch baseline}. It is also within 0.5\% of multi-epoch training on ResNet-50, indicating that hybrid training is a good default method to use for all architectures. 

\section{Conclusion}
Our work provides a new understanding of non-contrastive SSL methods like SimSiam and BYOL. We find that tricks like EMA, while unnecessary on ImageNet when using large models like ResNet-50, are indeed important when using smaller models on complex datasets. We show that the ratio between model capacity and dataset complexity determines when collapse occurs, and that smaller models like ResNet-18 and ResNet-34 suffer from increasing amounts of collapse when trained on larger subsets of ImageNet. We also show that increasing model width is better at improving performance than increasing depth.

We show that the singular values of the representations provide an effective metric to measure dimensional collapse. Furthermore, we find that a simple linear function of the validation loss and the AUC is highly predictive of the downstream linear probe accuracy. This relationship lets practitioners decide which model to use, without needing to obtain labels or do additional fine-tuning. 

Finally, we show that switching to a continual learning setting can prevent collapse by presenting manageable chunks of data to fit in sequence. By doing continual learning for half of training, then finishing with multi-epoch training, we outperform the vanilla SimSiam ResNet-18 by 18.3 percentage points. This also outperforms the BYOL ResNet-18 by 1.1 percentage points. Further work is required to fully understand why continual learning is helpful in this setting.\\

\noindent \textbf{Acknowledgements} We thank Sudeep Dasari for helpful discussions, Shikhar Bahl and Russell Mendonca for writing help, and Yufei Ye for valuable feedback. AL is supported by the NSF GRFP under grants DGE1745016 and DGE2140739. This work is supported by NSF IIS-2024594 and ONR N00014-22-1-2096.

\bibliographystyle{splncs04}
\bibliography{main}

\clearpage
\appendix
\section{Additional Experiments}
\subsection{ResNet-34 Results}
\begin{figure}
    \centering
    \includegraphics[width=0.4\linewidth]{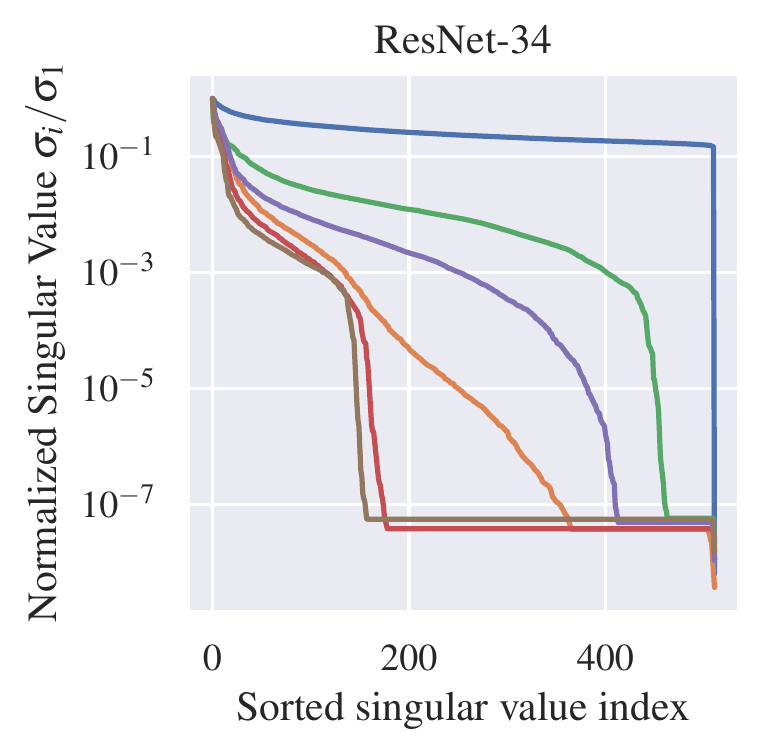}
    \includegraphics[width=0.4\linewidth]{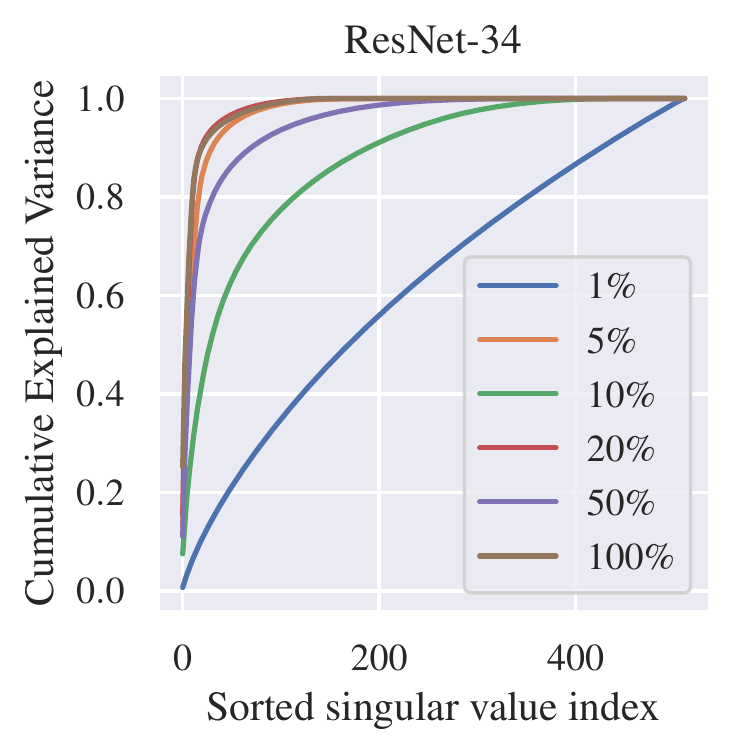}
    \caption{\textbf{Partial dimensional collapse for large subsets.} Just as we did in Figure 3 for ResNet-18, we show the singular values of the representations computed by ResNet-34 models trained on different size subsets of ImageNet. The model trained on 1\% exhibits no collapse, whereas more data (other than the 10\% model) tends to lead to more collapse. Note that the degree of collapse for ResNet-34 is in general much worse than it is for ResNet-18. We hypothesize that this is due to its increased depth but equivalent width, which makes it easier for SimSiam to lose information at every layer and compute collapsed representations.}
    \label{fig:rn34_dimensional_collapse}
\end{figure}

\subsection{Places365}
\label{subsec:places}

To test the generality of our results on ImageNet, we additionally examine the effect of model capacity and data order on the $256 \times 256$ version of Places365-Standard, which consists of 1,803,460 images of scenes from 365 categories \cite{zhou2017places}. These experiments use the same hyperparameters and augmentation strategy described in Section~\ref{subsec:experimental_setup}. However, note that the augmentation strategy was designed to maximize feature learning from ImageNet's object-centric images, which requires different priors on invariance than the scenes in Places365. Thus, this augmentation strategy likely is suboptimal for pretraining on Places. 
\begin{figure}
    \centering
    \includegraphics[width=0.49\linewidth]{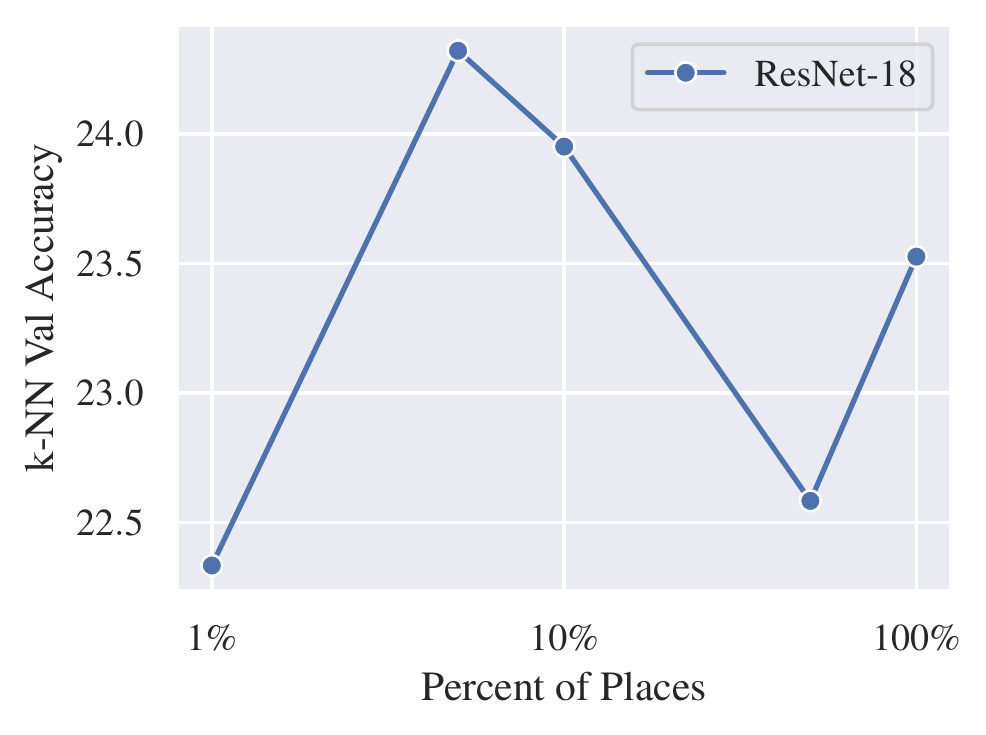}
    \includegraphics[width=0.49\linewidth]{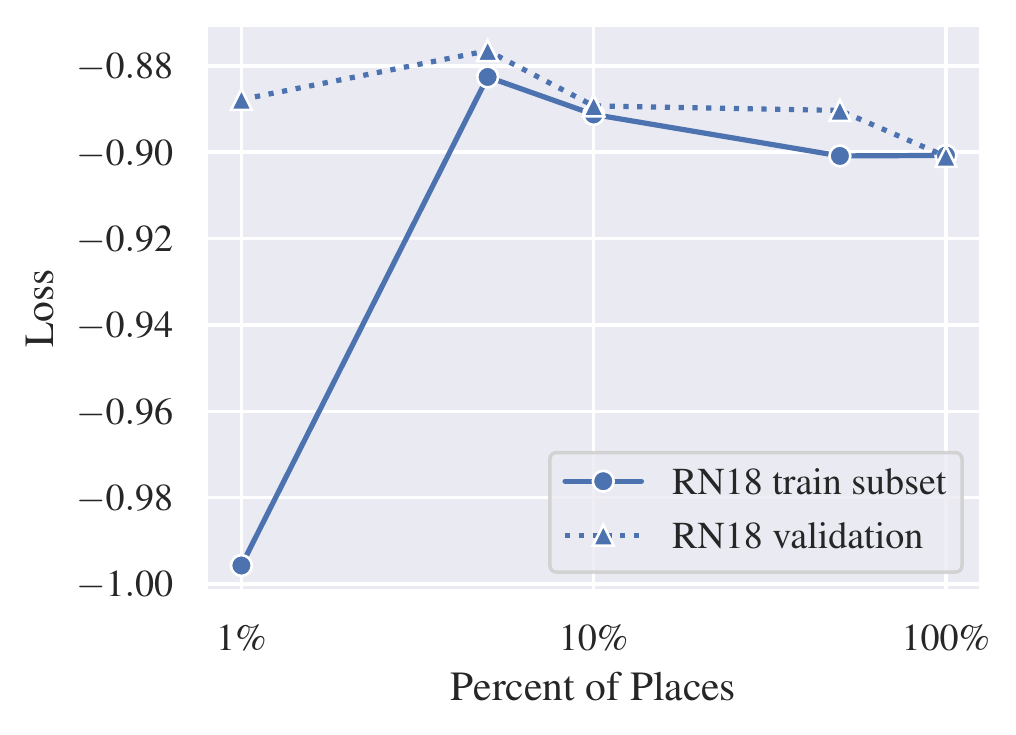}\\
    \caption{\textbf{Comparing models trained on subsets of Places365.} We trained 5 SimSiam models from scratch on Places365, each on a different size subset (1\%, 5\%, 10\%, 50\%, 100\%). Similar to what we found on ImageNet, the model trained on just a small subset of the data has the highest k-NN accuracy. This model (which uses 5\% of the training set) does not have the highest generalization ability. In fact, it achieves the worst validation loss out of the 5 trained models.}
    \label{fig:places}
\end{figure}

\begin{figure}
    \centering
    \includegraphics[width=0.49\linewidth]{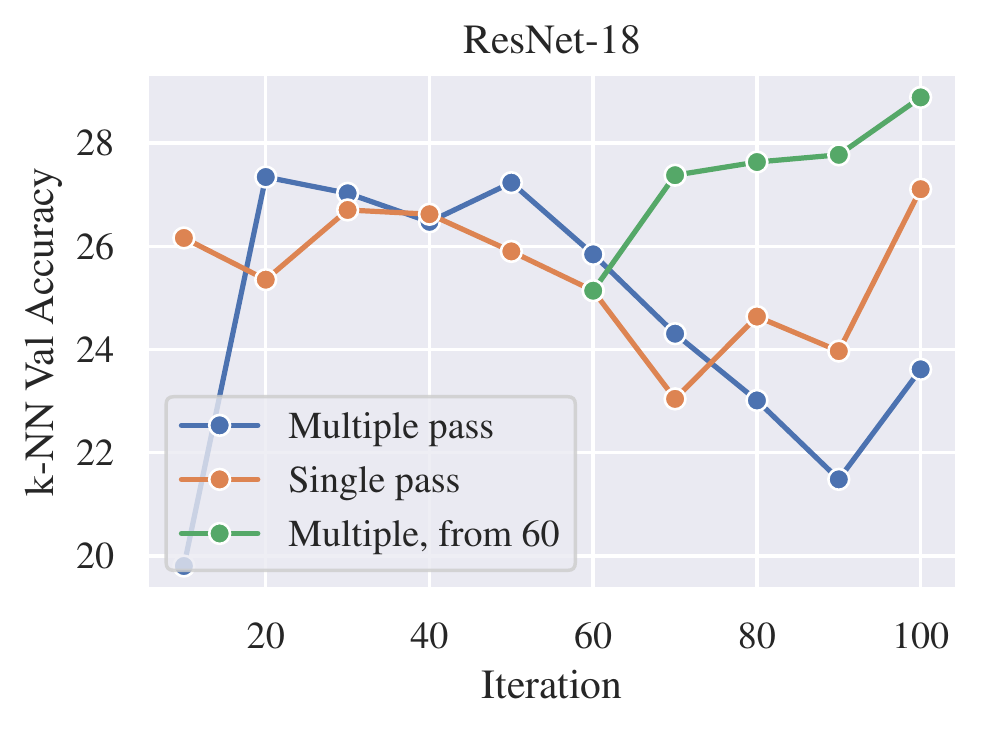}
    \includegraphics[width=0.49\linewidth]{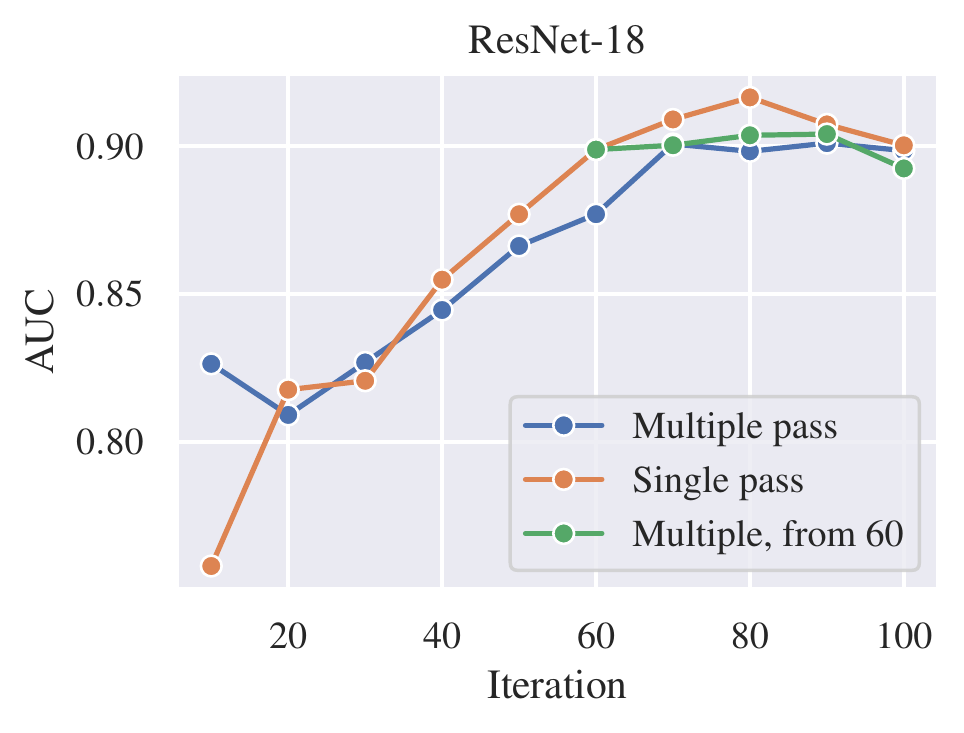}
    \caption{\textbf{Comparing intermediate checkpoints on Places365.} Unlike ImageNet, Places365 models tend to collapse much more over the course of training, as shown by the AUC plot on the right (higher implies more collapse). This may be due to the mismatch between the augmentation strategy and the scenes. This causes the ``multiple pass'' model accuracy to decrease significantly as it trains. However, hybrid training reduces collapse near the end of training and achieves the best accuracy out of all three data ordering methods. }
    \label{fig:places_across_training}
\end{figure}

\begin{table}
    \centering
    \caption{Places365 top-1 linear probing validation accuracy for different SimSiam training methods.}
    \begin{tabular}{lr@{\hspace*{4mm}}r@{\hspace*{4mm}}r}
    \toprule
        Training method & ResNet-18 \\
    \midrule
        Mutiple pass & 30.0 \\
        Single pass  & 33.6 \\
        Hybrid (switch at 60) & $\mathbf{35.0}$ \\
    \bottomrule
    \end{tabular}
    \label{tab:continual_places}
\end{table}

Just as we did in Section~\ref{subsec:dataset_size}, we trained SimSiam ResNet-18 models from scratch on Places365 using different size subsets of the training dataset (1\%, 5\%, 10\%, 50\%, 100\%). We find that the linear probing accuracy peaks when training on a tiny subset of the training set. Again, this performance is not explained by the training or validation loss. In fact, the best model achieves the worst loss. This result matches what we found on ImageNet in Figure~\ref{fig:dataset_size}. 

We also compared the different data ordering methods from Section~\ref{sec:continual}: multiple passes, single pass, and hybrid training. We choose to start hybrid training from epoch 60 instead of epoch 40, since Places365 has 50\% more images than ImageNet and we wanted to match the number of gradient steps taken using the multi-epoch phase across datasets. Table~\ref{tab:continual_places} shows that both single pass and hybrid training improve over standard multi-epoch training, and that hybrid training is the best of the three. We analyze the learning dynamics of these three orderings on Places365 in Figure~\ref{fig:places_across_training}. Training on Places365 is much more unstable, as the accuracy and collapse metric oscillate significantly across iterations. This may be due to the mismatch between the object-centric augmentation strategy and the structure of the Places365 scenes. The multiple pass model peaks early in training, around iteration 20, before collapsing over the next 80 iterations. In contrast, single pass training initially collapses more but recovers by the end of training, and hybrid training improves over the last 40 epochs. Overall, these results corroborate our hypothesis that collapse is responsible for decreased performance with smaller models and that continual or hybrid training is a viable solution to reduce collapse.

\subsection{Distillation}
\begin{table}[h]
\centering
\caption{Distillation yields strong performance using smaller networks, regardless of pretraining algorithm. We show the linear probe accuracy of the ResNet-50 teachers trained by SimSiam or MoCo-v3 and the student networks obtained by distilling each teacher network.}
\begin{tabular}{@{\extracolsep{4pt}}lccc@{}}
\toprule
\multirow{2}{*}{\textbf{Pretraining Alg.}}
&\multicolumn{1}{c}{\textbf{Teacher Net}} 
&\multicolumn{2}{c}{\textbf{Student Net}} \\
\cmidrule{2-2} \cmidrule{3-4}
& ResNet-50  & ResNet-18  & ResNet-34  \\
\midrule
SimSiam & 68.1 & 58.9 & 62.5 \\
MoCo-v3 & 66.4 & 56.4 & 57.5 \\
\bottomrule
\end{tabular}
\label{tab:distillation}
\end{table}

Model distillation \cite{hinton2015distilling,bucilua2006model} is a technique for compressing the knowledge in a large model into a smaller model. During the distillation process, the smaller student network learns to match the outputs of the larger teacher network. Training a large teacher model on a dataset (e.g. with cross-entropy loss) and distilling it into a smaller model typically performs better than directly training the small model. If we have a lot of compute available and only care about obtaining good small models, training a ResNet-50 with SimSiam (which does not collapse) and distilling it into a smaller model is an effective alternative approach for preventing partial dimensional collapse.

We distill a ResNet-50 into ResNet-18 and ResNet-34 by adding a fully connected layer that predicts the 2048-dimensional ResNet-50 representation and minimizes the mean-squared error (MSE). Note that this is equivalent to learning the top singular vectors of the teacher network representations, as the student network tries to learn a low-rank approximation of the teacher. Training the student is fairly straightforward. We train for 100 epochs on ImageNet-1k using the same hyperparameters used for SimSiam training. We find that the ResNet-50 teacher outputs are typically very small, on the order of 0.001 - 0.1, so minimizing the MSE with respect to the raw outputs leads to small gradient values and extremely long training times. Thus, we employ the standard trick of computing the mean and standard deviation of each dimension of the teacher output and using them to normalize the teacher output to have a mean of 0 and a variance of 1 in each dimension. We do this using an exponential moving average that updates the mean and standard deviation online. 

As expected, Table~\ref{tab:distillation} shows that distillation produces ResNet-18 and ResNet-34 networks with high linear probing accuracy. Note that this outcome is orthogonal to our work. First, distillation is incredibly compute-heavy. Training and distilling ResNet-50 into ResNet-18 takes as much as $4\times$ the compute as directly training the ResNet-18. Second, distillation is effective regardless of the pretraining algorithm -- MoCo and SimSiam both benefit from distillation. Finally, distillation performance does not resolve the fact that vanilla SimSiam uniquely has the partial dimensional collapse problem and can only be used to train smaller networks when using the data ordering strategies proposed in Section~\ref{sec:continual}.

\subsection{Vision Transformers}
\label{subsec:vit}
Self-supervised algorithms are typically evaluated using ResNets, but different architectures have qualitatively different behaviors. For example, self-supervised training with DINO \cite{caron2021emerging} leads Vision Transformers \cite{dosovitskiy2020image} to learn features corresponding to semantic segmentation, whereas ResNets trained with DINO do not. Thus, we experimented with using SimSiam to train Vision Transformers of varying sizes, in order to look for further architecture-related qualitative differences. Due to computational constraints, we tried ViT-Small, which was used in \cite{chen2021empirical}, as well as variants with fewer layers or attention heads. We trained these models using SimSiam for 100 epochs using the following hyperparameters from~\cite{chen2021empirical}: AdamW optimizer, learning rate of $1.5 \times 10^{-4}$, weight decay of 0.1, learning rate warmup for 10 epochs, and frozen linear patch projection. 

Surprisingly, these ViTs only achieve about 6-10\% linear probe accuracy. There could be several reasons for their poor performance. We could have used bad hyperparameters, although this indicates that SimSiam is very sensitive to hyperparameter values. This also could be due to their limited representation size (384 dim), which makes it less likely that they have learned many useful features. This is consistent with our findings in Section~\ref{subsec:dataset_size}. Finally, this could indicate that ViT fundamentally lacks some architectural inductive bias that makes non-contrastive algorithms like SimSiam or BYOL work with ResNet. Further work in this area could be illuminating. 

\subsection{Nearest Neighbors SimSiam}
We test whether a queue-based nearest neighbors loss (NNSiam, \cite{dwibedi2021little}) improves SimSiam training for ResNet-18. Given a pair of augmentations $x_1$ and $x_2$, the NNSiam objective is use $x_1$ to predict the nearest neighbor of $x_2$'s projected representation in a MoCo-style queue. We train for 100 epochs on ImageNet with the same hyperparameters as vanilla SimSiam and a queue of length 25600. 

This achieves a linear probe accuracy of 34.4\% on ImageNet, which is better than the vanilla ``multiple pass'' baseline (30.0\%), but still vastly underperforms our proposed methods, including ``single pass`` (44.5\%) or hybrid training (48.3\%). 

\subsection{Additional Baseline: Learning Rate Warmup}

\begin{table}
    \centering
    \caption{Additional baseline for comparing ImageNet validation top-1 linear probe accuracy for different SimSiam training methods.}
    \begin{tabular}{lr@{\hspace*{4mm}}r@{\hspace*{4mm}}r}
    \toprule
        Training method & ResNet-18 & ResNet-34 & ResNet-50 \\
    \midrule
    Multiple Pass & 30.0 & 16.8 & \textbf{68.09} \\
    Multiple Pass + 10-epoch lr warmup & 28.4 & 35.8 & - \\
    Hybrid (switch at 40) & \textbf{48.3} & \textbf{50.3} & 67.6\\
    \bottomrule
    \end{tabular}
    \label{tab:lrwarmup}
\end{table}
Figure 1(b) showed that MoCo-v3 \cite{chen2021empirical} and BYOL \cite{grill2020bootstrap} achieve reasonable performance with ResNet-18, whereas SimSiam collapses. One potential source of this difference is the learning rate warmup: MoCo-v3 and BYOL both utilize a linear learning rate warmup over the first 10 epochs, whereas SimSiam uses no warmup. We add a 10-epoch learning rate warmup to SimSiam and show that this detail is not responsible for the huge deficit in SimSiam performance. Table \ref{tab:lrwarmup} shows that warmup decreases ResNet-18 performance from 30.01\% to 28.38\% but increases ResNet-34 performance from 16.83\% to 35.82\%. This still falls quite short of the performance of our hybrid between single-pass and multi-pass training, which outperforms this baseline by 20 percentage points (ResNet-18) and 15 percentage points (ResNet-34). 
\end{document}